\pdfoutput=1
\documentclass[letterpaper]{article}
\usepackage{aaai18}
\usepackage{times}
\usepackage{helvet}
\usepackage{courier}
\usepackage[english]{babel}
\usepackage[utf8x]{inputenc}
\usepackage{amsmath}
\usepackage{amsfonts}
\usepackage{graphicx}
\usepackage{listings}
\usepackage{textcomp}
\usepackage{wrapfig}
\usepackage[dvipsnames]{xcolor}
\usepackage[hyphens]{url}
\usepackage[colorinlistoftodos]{todonotes}
\usepackage{microtype}
\usepackage{multirow}
\begin{document}
\title{Generating Triples with Adversarial Networks for Scene Graph Construction}
\author{Matthew Klawonn\\ Rensselaer Polytechnic Institute \\ Dept. of Computer Science \\ Troy, NY 12180 \\ klawom@rpi.edu \And Eric Heim \\ Air Force Research Laboratory \\ Information Directorate \\  Rome, NY 13441 \\ eric.heim.1@us.af.mil
}
\maketitle
\begin{abstract}
Driven by successes in deep learning, computer vision research has begun to move beyond object detection and image classification to more sophisticated tasks like image captioning or visual question answering. Motivating such endeavors is the desire for models to capture not only objects present in an image, but more fine-grained aspects of a scene such as relationships between objects and their attributes. Scene graphs provide a formal construct for capturing these aspects of an image.  Despite this, there have been only a few recent efforts to generate scene graphs from imagery.  Previous works limit themselves to settings where bounding box information is available at train time and do not attempt to generate scene graphs with attributes. In this paper we propose a method, based on recent advancements in Generative Adversarial Networks, to overcome these deficiencies.  We take the approach of first generating small subgraphs, each describing a single statement about a scene from a specific region of the input image chosen using an attention mechanism. By doing so, our method is able to produce portions of the scene graphs with attribute information without the need for bounding box labels. Then, the complete scene graph is constructed from these subgraphs.  We show that our model improves upon prior work in scene graph generation on state-of-the-art data sets and accepted metrics.  Further, we demonstrate that our model is capable of handling a larger vocabulary size than prior work has attempted.
\end{abstract}

Learning representations of visual scenes remains an important task that underlies many computer vision problems ranging from visual question answering \cite{malinowski2014towards} to image retrieval \cite{johnson2015image}. In order to be successful in these tasks, images must be represented in a form that captures details of the objects contained in a scene, including what objects are present, what attributes each object possesses, and how objects relate to one another. Much of the recent work on learning how to visually perceive images has focused largely on object detection and classification \cite{he2015deep,szegedy2016inception}.  These tasks focus on identifying one or more concepts or objects depicted in an image, but cannot produce representations that capture more complex characteristics or relationships of objects within scene.  Such information may provide insight necessary to understand a scene.

In this work, we focus on the task of learning to produce structured representations of images  that express rich scene information. More specifically, our goal is to learn a model that is able to generate a \emph{scene graph} \cite{johnson2015image} given an image. A scene graph describes the content of a scene by representing objects within an image as nodes and relationships between objects as edges. By including nodes that correspond to visual properties, an object can be represented as having an attribute if an edge is drawn between said object and attribute. As such, scene graphs are naturally able to model not only what objects are in a scene, but how they relate to each other and what attributes they possess. 
{
\centering
\begin{figure}
\includegraphics[]{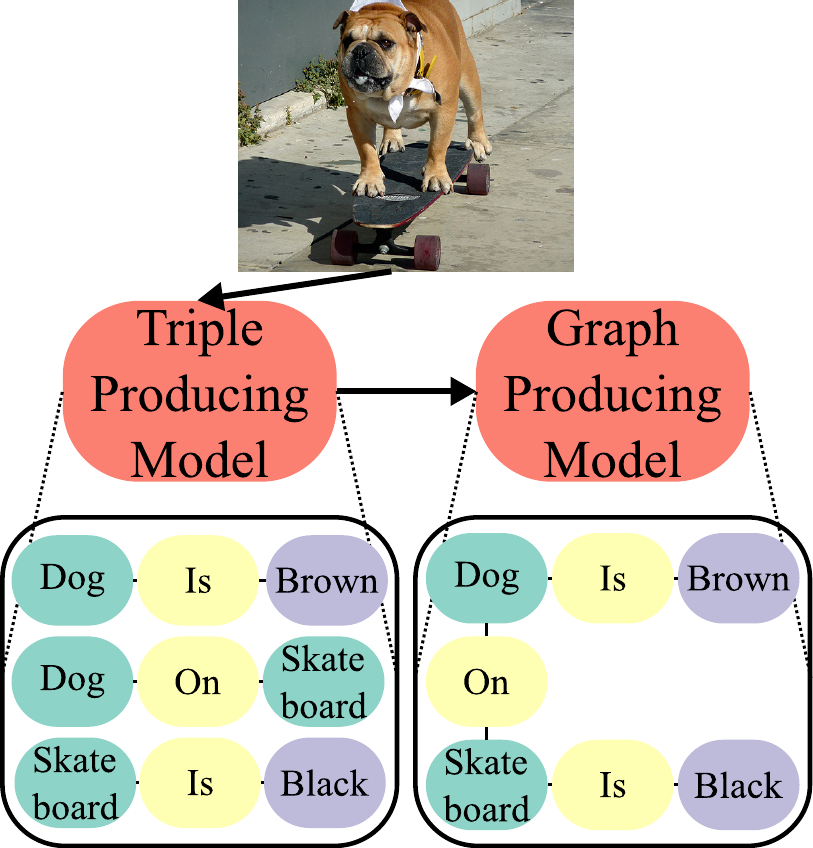}

\caption{Depiction of our model.  Given an image of a scene (top), our model (middle-left) generates triples (bottom-left) that express statements about objects in the image.  Then, duplicate entities in the triples with the same lexeme are determined by examining if they are from similar regions of the image (middle-right).  These duplicates are merged into single nodes to form a scene graph (bottom-right).}
\label{fig1}
\end{figure}
}

For a learned model to map an image to a scene graph, it must both have the capacity to represent high-dimensional inputs and the ability to produce complex, structured outputs. Generative Adversarial Networks (GANs) \cite{goodfellow2014generative} have shown success for such learning problems. In the GAN framework a \emph{generator} neural network is pitted against an adversary  network called a \emph{discriminator}. The network are trained in tandem,  eventually reaching an equilibrium where the generator can produce output that is indistinguishable from real data by the discriminator. 

While prior GAN models have almost exclusively focused on generating images, we propose a novel GAN that is able to generate plausible scene graphs for given images.  Our generator produces individual statements about a scene called \emph{triples}, which are three lexeme sequences that describe object relationships or attributes.  Because the generator is trained to provide a variety of outputs given an image, our generator is able to produce many different triples, each putting forth a different statement about a scene.  To combine the disparate triples into a proper scene graph we utilize an attention mechanism \cite{xu2015show} that is trained within our generator to determine if two lexemes with the same label were produced from the same spatial region of the input.  If they are, we merge the two lexemes in our scene graph (see: Figure \ref{fig1}).  

In summary the contributions of this work are as follows:

\begin{enumerate}
\item We formulate a novel GAN model that is able to generate triples over a scene that contain both object relationship and attribute statements.
\item We use an attention mechanism both to generate parts of a scene graph from different parts of an image, and to merge these subgraphs together.
\item We demonstrate empirically that our model improves on the state of the art in scene graph generation while also being able to model object attributes.
\end{enumerate}

The remainder of this paper is organized as follows. We first discuss previous, related work and compare our method to other approaches for scene graph generation. We then discuss our proposed approach, beginning with a formal definition of our learning problem.   Next, we quantitatively evaluate our method against the state-of-the-art in scene graph generation, and qualitatively discuss illustrative examples of generated scene graphs.  Finally, we conclude and discuss directions of future work.

\section{Related Work}
Work related to ours lies primarily in three areas: methods for learning representations of visual scenes, Recurrent Neural Networks (RNNs), and GANs.  Next, we review these areas in relation to our work. 

\subsection{Scene Understanding and Representation}
A simple way to describe a scene is to list the objects it contains.  This task is often called visual object recognition, a classification task for which numerous leaps in performance have been achieved over the last half-decade~\cite{krizhevsky2012imagenet,simonyan2014very,he2015deep,szegedy2016inception} on large-scale image data sets, such as ImageNet~\cite{russakovsky2014imagenet}.

More sophisticated approaches to describing a scene, like image captioning \cite{vinyals2014show}, dense captioning \cite{johnson2015densecap}, and visual question answering (VQA) \cite{malinowski2014towards} provide more expressive means of capturing the details of an image  than simply listing objects. Each task involves producing natural language output given an image as input. While these representations have been shown to be helpful for various tasks (e.g image retrieval), the outputs can be interpreted in many different ways due to the ambiguity of natural language. Further, these tasks do not explicitly attempt to describe the entirety of a scene or link information from various parts of a scene together.

One way to limit the ambiguity of natural language and explicitly model object relationships is to describe scenes using scene graphs, first proposed by~\cite{johnson2015image}. Scene graphs have been shown to be useful in a number of tasks like content based image retrieval~\cite{johnson2015image} and automatic caption evaluation~\cite{anderson2016spice}. In one of the first works in generating scene graphs, the authors of ~\cite{anderson2016spice} propose a method to generate scene graphs from captions. More recently,  the authors of ~\cite{xu2017scenegraph} propose a method to construct a scene graph from an image by first fixing the structure of the graph, then refining node and edge labels using iterative message passing. Their work limits itself to training with the use of bounding boxes and doesn't predict attributes, likely because multiple attributes may share a bounding box with a single object, and learning to generate multiple attributes given a single bounding box is non-trivial. In contrast, our method learns to generate individual triples, using the GAN training framework coupled with attention to focus on regions.  As such, our model does not require bounding box labels to train and is not limited to relating two separate objects that were labeled in the image, thus enabling attribute information to be generated.

\subsection{Recurrent Neural Networks}
Our architecture contains a recurrent language component to produce triples.  More specifically, we utilize Long Short Term Memory (LSTM)~ \cite{hochreiter1997long} networks with attention. The idea of an attention mechanism, first proposed in~ \cite{bahdanau2014neural}, is to allow a recurrent network access to the entire input at every timestep, with the attention mechanism determining what parts are important. Typically, an attention mechanism is simply a multilayer perceptron jointly trained with the rest of the recurrent architecture. \cite{xu2015show} are the first to use an attention mechanism in an LSTM with an image as input, with many others following~\cite{yang2015stacked,shih2015look,xu2015ask,lu2016hierarchical}. Importantly, we explicitly use the attention mechanism to disambiguate entities in produced triples, a technique that to our knowledge has not been applied in prior work. We use LSTMs because they are historically one of the most successful RNNs, and offer comparatively similar performance to more recent variants, such as GRUs \cite{chung2014empirical}.

\subsection{Generative Adversarial Networks}

Generative Adversarial Networks have generated interest since their inception in \cite{goodfellow2014generative} because of their empirical successes in modeling high dimensional data distributions. Recently, \cite{arjovsky2017towards} and~\cite{arjovsky2017wasserstein} illustrated a number of theoretical and practical issues with the original GAN formulation, and proposed to train the discriminator to measure the difference between generated and ground truth samples with a Wasserstein metric rather than a Kullback-Leibler divergence, accomplished by a change in loss functions. In order to generate sequences using one hot vector ground truth examples and continuous vector based generated outputs, this change is necessary. As explained in \cite{gulrajani2017improved}, the KL divergence between two one hot vectors is infinite resulting in useless gradients. 

While applications of GANs have largely been limited to generating images~\cite{radford2015unsupervised,ledig2016photo,reed2016generative}, there have recently been a few efforts to generate natural language. Such endeavors use training methods other than backpropagation~\cite{yu2016sequence}, or a continuous approximation to sampling from a categorical distribution~\cite{jang2016categorical,maddison2016concrete}. We leave the training of scene graph generation models with these approaches for future work. There have also been a number of works in using GANs with an input condition, e.g generating images conditioned on text \cite{reed2016generative,han2017stackgan} or other images \cite{ledig2016photo}. Like previous work, we condition our GAN on images;  unlike previous work, our output is triples.

\section{Generating Scene Graphs from Imagery}
In this section, we formalize the problem of generating a scene graph and discuss the motivation and specifics of our approach. A unique characteristic of our model is that scene graphs can be produced by generating their subunits and stitching them together. As such, we outline architectures and training approaches suitable for this formulation of the problem, specifically motivating the use of attention mechanisms and GANs. Finally, we provide implementation details.

\subsection{Problem Formulation}
The goal of this work is to find a mapping between an image of a scene and a scene graph. Specifically we propose to learn a mapping $g_{\Theta}: I \rightarrow G$ from a color image $I\subset \mathbb{R}^{d_1\times d_2\times 3}$ to a scene graph $G = \left(V,E\right)$. Each vertex $v\in V$ and edge $e=\left(v_1^e, v_2^e\right)$ is labeled by a lexeme in a vocabulary $\mathcal{V}$.

Each edge in a scene graph defines a single statement about a scene called a \emph{triple}.  Let $l_e$ be the label for edge $e$ and $l_v$ be the edge for vertex $v$.  Every edge $e =\left(v_1,v_2\right) \in E$ can be formatted as a triple $t_e = \left(l_{v_1},l_e,l_{v_2}\right)$.  We call these triples because they are similar in spirit to Resource Description Framework (RDF) triples \cite{lassila1999resource},  which are three entities representing some statement about data in the form of a subject-predicate-object structure.  In the scene graphs considered in this work, triples either describe \emph{relations} between two objects in the scene (e.g. $t_e$ = (``dog'',``on'',``skateboard'') in Figure \ref{fig1}), or state that an object has an \emph{attribute} \cite{ferrari2008learning}, a mid-level  visual  concept  (e.g. $t_e$ = (``dog'', ``is'', ``brown'') in Figure \ref{fig1}).

With these definitions, two things are clear. Individual triples are capable of describing the contents of a scene, while arranging them into a graph resolves duplicate entities into a single vertex. Seeing that these two steps can be separated, we map images to scene graphs by first generating triples given an image, and then resolve objects that are the same to construct a proper scene graph. We hypothesize that by focusing on the quality of generated statements, saving the structuring of the graph for later, we can more accurately predict components of a scene graph. More formally, we choose to first find a mapping $g'_\Theta : I \mapsto \mathcal{V} \times \mathcal{V} \times \mathcal{V}$ to find triples over a scene, and then a mapping $g'' : \left\lbrace\mathcal{V} \times \mathcal{V} \times \mathcal{V} \right\rbrace^n\mapsto G$ to then produce a scene graph from the triples.  First we discuss our method for learning $g'_{\Theta_g}$.

\subsection{Triple Generation Network}
For the generation of triples, we propose the a neural network architecture based on ideas from Convolutional Neural Networks (CNNs), RNNs, and GANS.  The network has two components, the first feeding into the second.   The first is a \emph{feature extractor} $f_{\Theta_f} : I \mapsto I'$ that maps images to visual features using a fully convolutional neural network.  There are $L$ total features produced, and each  belongs to $\mathbb{R}^D$, with $D$ corresponding to the number of convolutional filters in $f_{\Theta_f}$. These features are fed into a \emph{recurrent} component $r_{\Theta_r} : I' \mapsto \mathcal{V} \times \mathcal{V} \times \mathcal{V}$.  More specifically, we utilize a Long Short Term Network (LSTM) that outputs a lexeme from the vocabulary for each of three time steps.   Given visual features $\mathbf{X}'_t \in I'$, an LSTM unit at time $t$ is defined by the following: 
\begin{equation} \label{eq1}
\begin{split}
\vec{f}_t & = \sigma\left(W_f \mathbf{X}'_t + U_f \vec{h}_{t-1} + b_f \right) \\
\vec{i}_t & = \sigma\left(W_i \mathbf{X}'_t + U_i \vec{h}_{t-1} + b_i \right) \\
\vec{o}_t & = \sigma\left(W_o \mathbf{X}'_t + U_o \vec{h}_{t-1} + b_o \right) \\
\vec{c}_t & =  f_t \circ \vec{c}_{t-1} + \vec{i}_t \circ tanh \left(W_c \mathbf{X}'_t + U_c \vec{h}_{t-1} + b_c\right)\\
\vec{h}_t & =  o_t \circ tanh(\vec{c}_t)
\end{split}
\end{equation}
The hidden state $c_t$ is used to aggregate information from the previous step's LSTM output $h_{t-1}$ and the input $\mathbf{X}'_t$. The vectors $\vec{f}$, $\vec{i}$, and $\vec{o}$ are \emph{gates} that determine what information from $\mathbf{X}'_t$ (visual features), $\vec{h}_{t-1}$ (output of previous time step), and $\vec{c}_{t-1}$ (previous hidden state) is and is not used to determine the current hidden state $\vec{c}_t$. Intuitively, these gates are attempting to explicitly learn what to ``remember'' and what to ``forget'' so that long-term dependencies between inputs and outputs can be modeled. The vector $\vec{h}_t$ is the output produced by the LSTM at step $t$ based on $\vec{c}_t$. In order to map $\vec{h}_t$ to lexemes in our vocabulary we learn an additional affine transformation of $\vec{h}_t$ to produce a $|\mathcal{V}|$ length vector, where each element corresponds to a lexeme in $\mathcal{V}$. This vector is then normalized so the elements sum to one, and the index of the highest value can be used index the vocabulary to finally output a human readable label.  This is done at all three time steps of the LSTM to create a triple.
Note, as currently defined, the sole input to the recurrent component is the output of the convolutional component.  As such, only one triple will be generated per input image.  Because a scene can contain many different objects with many different relations and attributes, we want  our generator to model a distribution over triples, given an image.  Motivated by recent successes in learning generative distributions over structured data using GANs, we train our generator using an adversarial strategy that enables us to sample a variety of triples to use as a basis for scene graph construction. 

\subsection{Adversarially Training the Triple Generator}
To train $g'_{\Theta_g}$ we pit it against a discriminator $d_{\Theta_d} : \left(V \times V \times V\right) \times I' \mapsto \left[0,1\right]$, where $V = \left[0,1\right]^{|\mathcal{V}|}$.  The  goal of $d_{\Theta_d}$  is to discriminate ground truth triples, encoded as three one-hot vectors in a training set, from three outputs of  $g'_\Theta$, given an image.  The generator and the discriminator are trained in tandem, in an adversarial game.  The recurrent component of our generator takes visual features of an image $\mathbf{X}'$ concatenated with a random vector $\vec{n} \sim \mathcal{N}\left(0,\mathbf{I}\right)$, and produces a triple $\tilde{t}_e = \left( v_1, v_2, v_3\right)$.  The random noise portion of the input ensures $g'_\Theta$ does not deterministically produce a single unique triple given $\mathbf{X}'$, and can instead produce a variety of triples.  Concatenation of the $\mathbf{X}'$ onto $\vec{n}$ can be viewed as conditioning if the generator is viewed as a probability distribution.  We adopt notation to reflect this view that  $g'_\Theta$ and $d_{\Theta_d}$  are probability distributions conditioned on an image.  From the training set, a ground truth triple $t_e$ associated with $\mathbf{X}'$ is sampled. The discriminator is trained to output a low value when given $(\tilde{t}_e \times \mathbf{X}')$, and a high value when given $(t_e \times \mathbf{X}')$, separating the real triples from generated ones as much as its architecture allows. Training is performed end to end using stochastic gradient descent with backpropagation. As a result error information from the discriminator is propagated to the generator, informing it how to generate triples similar to those in the training set.

In order for $d_{\Theta_d}$ to propagate error to $g_{\Theta_g}'$, it requires an appropriate loss function.  Towards this end, we use the Wasserstien metric to measure the distance between the data and generator distributions, a technique first introduced in~\cite{arjovsky2017wasserstein}.  This metric induces the following loss function:
\begin{equation} \label{eq2}
\mathcal{L}\left(\mathbf{X}',\vec{n};\Theta_g, \Theta_d \right) = d_{\Theta_d}\left(g'_{\Theta_g}\left( \vec{n}|\mathbf{X}'\right)|\mathbf{X}'\right) - d_{\Theta_d}(t_e | \mathbf{X}' ) 
\end{equation}
Minimizing this loss with respect to $\Theta_d$, allows the $d_{\Theta_d}$ to better discriminate samples from $g_{\Theta'_g}$ from those from the ground truth.  Maximizing the loss with respect to $\Theta_g$ allows $g_{\Theta'_g}$ to produce samples that ``fool" the discriminator into producing high scalar valued scores for generated triples, indicating the samples look like ground truth samples.  As such, our model is trained in two phases.  First, $d_{\Theta_d}$ is updated by sampling $\mathbf{X}'$ and $\vec{n}$ and performing a stochastic gradient minimization update with respect to $\Theta_d$.  Then, $g_{\Theta_g}'$ is updated by sampling $\mathbf{X}'$ and $\vec{n}$ and performing a stochastic gradient maximization update with respect to $\Theta_g$.

Because our training algorithm promotes sampling of different triples from a single image, we do not require human or machine provided bounding boxes, a significant benefit of our approach over prior work. Since $d'_{\Theta_d}$ alone informs our generator, the task of $g'_{\Theta_g}$ is not to predict a correct output given a bounding box offset as it is in \cite{xu2017scenegraph}, but rather to predict an output that the discriminator will score highly.  While this eliminates the need for bounding box information in generating triples, our method requires  $g''$ to construct a proper scene graph.  The challenge in this task is that separate triples may contain lexemes that refer to the same object.  This ambiguity must be resolved in order to successfully construct a scene graph from generated triples.  Our approach to resolve lexemes is to associate a separate spatial region of the input with each generated lexeme.  We do this using an attention mechanism that produces a vector that quantifies which regions of the image most influenced the generation of the lexeme.  If two of the of same generated lexemes have similar attention vectors, we can determine that they are the same object in the scene.  We outline this process further in the next section.

\subsection{Graph Construction from Attention}
An attention mechanism is a differentiable function of trainable parameters that accepts as input some collection of features and outputs a relative importance of said features. We utilize the mechanism introduced in \cite{xu2015show} in the recurrent component of our generator.  This mechanism is defined as  follows:
\begin{equation} \label{eq3}
\begin{split}
e_{ti} & = a_{\Theta_a}(\vec{x}'_i, \vec{h}_{t-1}) \\
\alpha_{ti} & = \frac{\exp(e_{ti})}{\sum_{k=1}^L \exp(e_{tk})}\\
\vec{z}_t &= \sum_{i=1}^{L}\alpha_{ti}\vec{x}'_i
\end{split}
\end{equation}
Here, $a_{\Theta_a}$ is a multilayer perceptron that accepts a concatenation of the input convolutional features $\vec{x}'_i$ and the hidden state of the previous step $\vec{h}_{t-1}$ in our recurrent component, and produces a vector $\vec{e}_t$ for which each element is a relative importance of each visual feature in $\mathbf{X}'_t$.  This vector is then normalized with a softmax to form $\vec{\alpha}_t$.  Finally, the elements of $\vec{\alpha}_t$ are used to weigh the $L$ visual features to produce a vector $\vec{z}_t \in \mathbb{R}^D$ which is used as input to our recurrent component $r_{\Theta_r}$ instead of the using all $L$ visual features in $\mathbf{X}_t'$.

Because $\vec{\alpha}_t$ gives a relative importance of the visual features input to $r_{\Theta_r}$ at time $t$, and because these visual features are the product of a fully convolutional neural network $f_{\Theta_f}$, $\vec{\alpha}_t$ gives an explicit weighting of the spatial regions in the image used to generate a particular output. As a result each of the three lexemes in each triple generated by our architecture can be mapped to a region or regions of the input image.  Thus, our recurrent component generates lexemes based on specific, emphasized regions of the image. 

{
\centering
\begin{figure*}[t]
\includegraphics[]{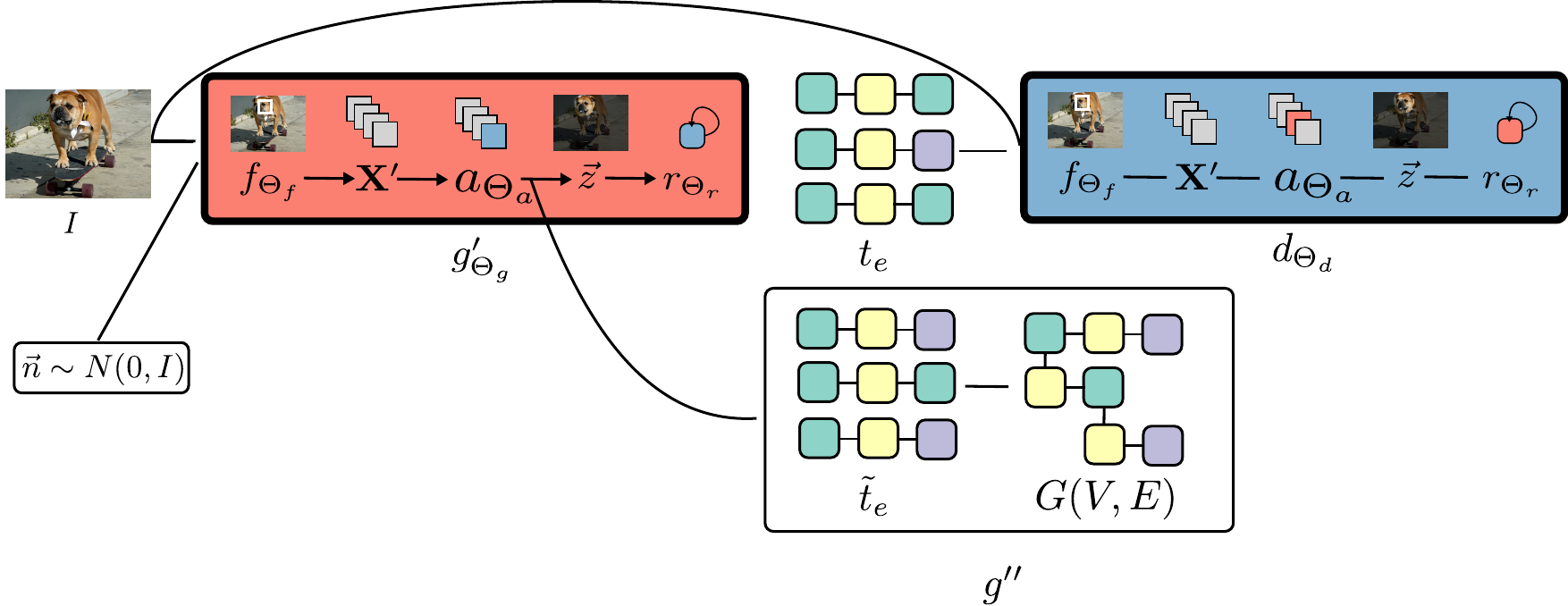}

\caption{Our architecture: Input images $I$ and noise $\vec{n}$ are first fed to the generator $g_{\Theta_g}$ (red), which processes the image using a CNN $f_{\Theta_f}$, generates image features $\mathbf{X}'$, passes those to an attention mechanism $a_{\Theta_a}$that generates a dynamic image representation $\vec{z}$ and attention vector $\vec{\alpha}$. $\vec{z}$ is fed to an LSTM that produces triples $\tilde{t}_e$. During training, these triples are passed along with ground truth triples $t_e$ to the discriminator $d_{\Theta_d}$ (blue) that contains the same components as the generator. The discriminator only produces a score, however. During test time all $\tilde{t}_e$s and $\alpha$s are passed to $g''$ which resolves the triples into a graph $G(V,E)$.}
\label{fig2}
\end{figure*}
\par}

Note that at the first step $t=1$ of LSTM calculation there is no value for $\vec{h}_{t-1}$ ($\vec{h}_0$), so in \cite{xu2015show} it is predicted as a learned affine transformation of the average of all $L$ features, the idea being to give the an "overview" of the content of the image to the attention mechanism. In contrast we would like to initialize this value such that the attention mechanism is conditioned on a random portion of the image to start rather than the whole image, increasing the variety of triples that can be produced. Thus we map a random $\vec{x}'_i \in \mathbb{R}^D$ to $h_0$ via a learned affine transformation. 

In order to resolve lexemes across triples, i.e to create our mapping $g''_\Theta$, we compare the  $\vec{\alpha}_t$ vectors of lexemes with the same label using  generalized Intersection over Union (IoU):

\begin{equation} \label{eq4}
\begin{split}
\mbox{IoU}(\vec{x}, \vec{y}) & = \frac{\sum_i \min \vec{x}_i \vec{y}_i}{\sum_i \max \vec{x}_i \vec{y}_i}
\end{split}
\end{equation}

\noindent If their similarity is over a threshold, then the two lexemes can be deemed one and the same.  This use of the attention vectors produced for each output is, to the best of our knowledge, a novel application of said vectors.
 
We now have all the components necessary to train our generator $g'_{\Theta_g}$and then construct scene graphs via $g''$. Our full architecture flow to produce one triple $\tilde{t}_e$ is as follows. First, we use $f_{\Theta_f} : I \mapsto I'$ to extract convolutional features $\mathbf{X}' \in \mathbb{R}^{D \times L}$, which are subsequently fed into our attention mechanism $a_{\Theta_a}$. The attention mechanism produces vectors $\vec{z}, \vec{\alpha} \in \mathbb{R}^D$. $\vec{z}$ is concatenated with $\vec{n}$ and fed to our LSTM which produces an output $\vec{h}_t$, then transformed via affine parameters to generate a single lexeme $\vec{v}_t = \left(W_v \vec{h}_t + b_v \right)$. We repeat this process three times to generate $\vec{v}_1$, $\vec{v}_2$, $\vec{v}_3$, resolving these lexemes to their labels to produce the triple $\tilde{t}_e = \left(l_{v_1},l_{v_2},l_{v_3}\right)$. A number of triples are then combined via their associated attention vectors $\vec{\alpha}$ to form a scene graph proper. With our full architecture outlined here and in Figure \ref{fig2}, we now provide an empirical evaluation and details on our implementation.

\section{Empirical Evaluation}
One goal of our evaluation is to compare our method  to the current state-of-the-art in scene graph generation. As \cite{xu2017scenegraph} sets the current state-of-the-art, we compare to their method, using metrics their work established, and on the dataset they evaluated on.  All data comes from the Visual Genome (VG) dataset \cite{krishna2016visual}, since this is the largest and highest quality dataset containing image-scene graph pairs available today and the same data that \cite{xu2017scenegraph} use for evaluation. In addition, to demonstrate our model's ability to generate attributes in addition to relations, we show the performance of our model trained on an extended vocabulary that contains attribute triples.

\subsection{Experimental Set-up}
For the feature extractor component $f_{\Theta_f}$, we use the convolutional layers of \cite{simonyan2014very} to produce input features, which are standardized. $f_{\Theta_f}$ is not trained with the rest of the architecture, but rather pre-trained on the image classification task of \cite{russakovsky2014imagenet}. Following the example of \cite{gulrajani2017improved}, we use the Adam stochastic gradient algorithm \cite{kingma2014adam} with learning rate $1e-4$, $\beta_1 = 0.5$, $\beta_2 = 0.9$ to train both the discriminator and generator.  Our gradient penalty coefficient $\lambda$ \cite{gulrajani2017improved} is set to 10, and no hyperparameter optimization is done. Each layer of our architecture uses layer normalization \cite{ba2016layer} to help avoid saturating LSTM non-linearities. For our graph construction phase, entities that have more than a 80\% match using the generalized IoU metric are considered to be duplicate entities. In the cases where more than one label appears for these entities, we use majority voting to determine a single label. We note that this threshold could be tuned on a validation set by some metric, e.g the recall @ k metric discussed in the next section.

VG contains over 108,000 image/scene graph pairs. We create two splits of these pairs. The first split exactly matches that of \cite{xu2017scenegraph}, which is a 70-30 train-test split of the dataset capturing the top 150 object classes and 50 relations for a vocabulary size of 200.  The second split is also a 70-30 train-test split, but includes the 400 most common object classes, 150 most common relations, and 150 most common attributes.  The first split allows us to compare our model directly to the results reported in \cite{xu2017scenegraph}.  The second not only expands the vocabulary to a larger number of relations, but also includes attributes, enabling evaluation of our model on a larger variety of possible triples and more complex scene graphs. 

In \cite{xu2017scenegraph}, there are three separate tasks that are used to evaluate their models: predicate classification, scene graph classification, and scene graph generation. In predicate classification, the task is to predict the correct predicate for a relation triple with the two objects being given, along with their bounding boxes. In scene graph classification, the task is to predict object labels and a predicate given a set of bounding boxes. In this work we only examine the task of scene graph generation as we do not use bounding box information, making the tasks of predicate and scene graph classification largely irrelevant for measuring our methods performance. 

The metric used in \cite{xu2017scenegraph}, recall at k, is defined as the fraction of the $k$ most likely generated triples that appear in the ground truth, relative to the total number of triples in the ground truth. For a set of ground truth triples $t_e$ and generated triples $\tilde{t}_e$ we formally define this in \eqref{eq5}:

\begin{equation} \label{eq5}
\begin{split}
r@k &= \frac{\left| \tilde{t_e} \cap t_e \right|}{\left| t_e \right|}\\
\end{split}
\end{equation}

The idea behind this metric is that models should not be penalized for making true predictions that don't happen to be in the ground truth data. Since labels are sparse and very few (if any) scenes are labeled with all object relationships and attributes, such a penalty would significantly cloud interpretation of results. When reporting results, we multiply the fraction given by $r@k$ by 100 to create a percentage.

Here we present our qualitative and quantitative results of our evaluation. In order to filter predictions to the top k, we generate a large sample of $\sim$ 500 triples, and use the discriminator to score each of the generated triples and rank by score from highest to lowest, with a high score indicating the discriminator thinks the triple is likely to come from the ground truth distribution. We refer to our method as SG-GAN, an abbreviation for Scene Graph GAN.

\subsection{Results}
\begin {table}[ht]
\caption {Recall @ 50 and 100 for the task of scene graph generation on the split of \protect\cite{xu2017scenegraph}. We compare only to their best results and nearly double the performance in both cases.} \label{tab:xu_split} 
\begin{center}
\begin{tabular}{ |c|c|c| } 

\hline
Metric & SG-GAN & \cite{xu2017scenegraph} \\
\hline 
r @ 50 & 6.84 & 3.44 \\
r @ 100 & 8.95 & 4.24 \\
\hline
\end{tabular}
\end{center}
\end{table}

\begin {table}[ht]
\caption {Recall @ 50 and 100 for the task of scene graph generation on our own split of both attributes and relations. There is no prior work in generating both using the same architecture and so we make no comparisons.} \label{tab:our_split} 
\begin{center}
\begin{tabular}{ |c|c|c| } 

\hline
Metric & SG-GAN & \protect\cite{xu2017scenegraph} \\
\hline 
r @ 50 & 1.74 & -- \\
r @ 100 & 2.47 & -- \\
\hline
\end{tabular}
\end{center}
\end{table}

Table \ref{tab:xu_split} shows the comparison between our model and the state of the art reported in \protect\cite{xu2017scenegraph}. On this form of the VG data, our method achieves approximately double the recall@50 and recall@100 as  \protect\cite{xu2017scenegraph}.  Again, we reiterate, this performance was achieved without the need for bounding box labels.  For our custom split of the VG data we observe a reduction in performance (Table \ref{tab:our_split}), when switching to the model trained and evaluated on our custom split of both relations and attributes.  

We speculate that the inclusion of attributes makes for a more difficult learning task than the increase in vocabulary alone for a few reasons. The visual cues necessary to detect relations and those necessary to detect attributes vary significantly, increasing the burden on any architecture attempting to capture both at the same time. Another reason is that the attention mechanism is likely to behave very differently depending on whether or not it is generating a relation vs an attribute. For generating an attribute one would expect the attention mechanism to look in the very near proximity of the object in contrast to all over the image. While neural networks have a very large capacity and ought to be able to ``remember" how to treat the two differently, we suspect the stark difference between looking in a different location for each lexeme produced and looking in the same location for each lexeme produced is still challenging.

While the performance gain is significant over prior work, the recall of our model is low relative to related visual inference tasks.  This is somewhat expected, given that scene graph generation combines a number of these tasks as sub-problems such as object detection, object/object relationship identification, and attribute prediction. As a result, we note that the task of scene graph generation remains a challenging problem which likely presents an opportunity for future work to improve on our results.

{
\centering
\begin{figure*}[t]
\includegraphics[]{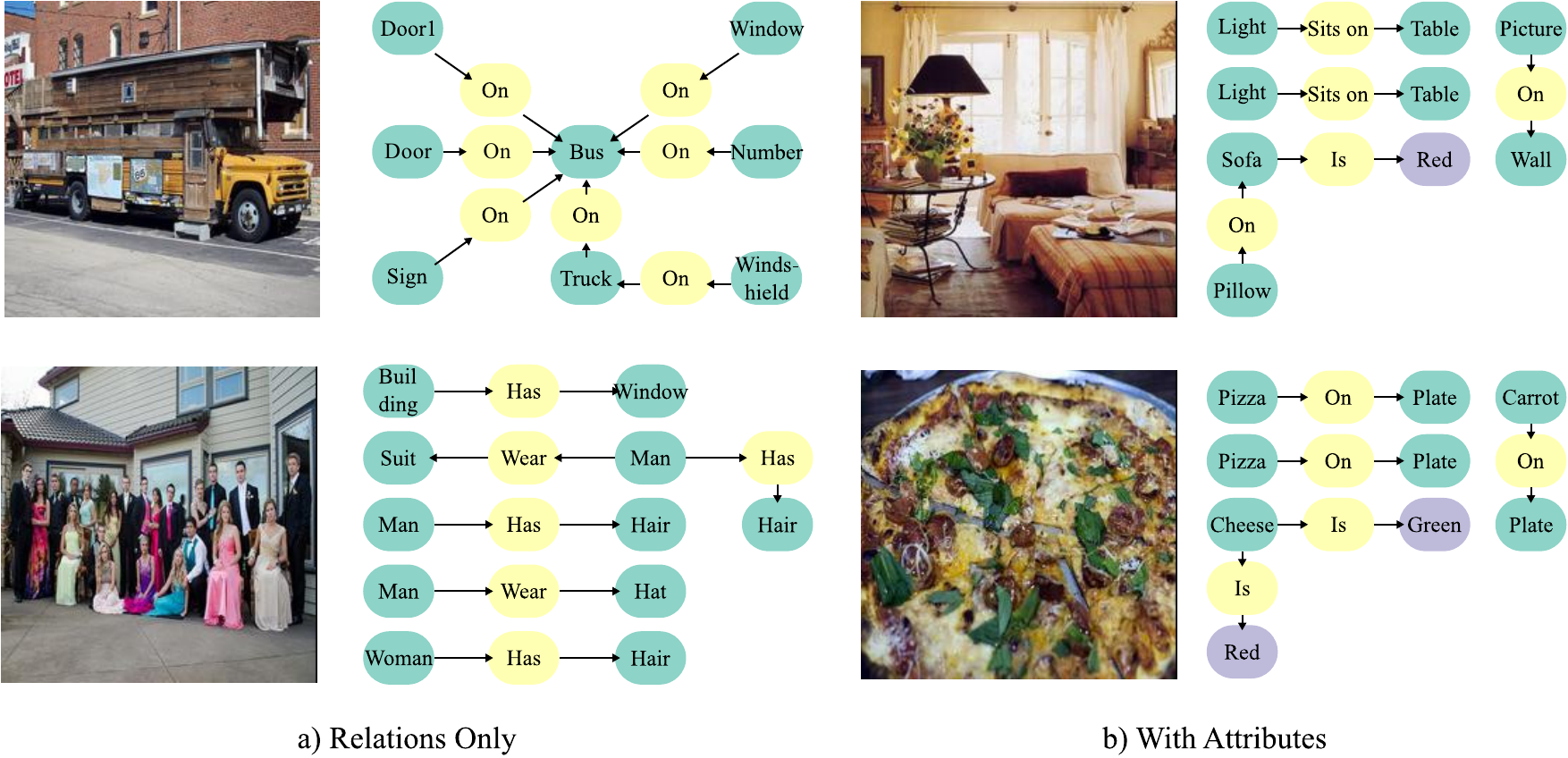}

\caption{Example scene graphs generated by SG-GAN. a) These scene graphs were generated by the model trained on the relations only data split. b) These were generated by the model trained on both relations and attributes.}
\label{fig3}
\end{figure*}
\par}

In Figure \ref{fig3} we show examples of scene graphs that we generated using our relations only model and our relations-attributes model respectively. For interpretability we limit the triples generated to the top 10 (or below if duplicate entities are resolved) as determined by our discriminators. The results are not out of line with how we train our model. Many of the triples that exist in the graphs generated by our method are either true or close to being true. In particular object detection seems to work fairly well, which is also unsurprising given that \cite{simonyan2014very}, the pretrained model that provides our convolutional features, is trained on an object detection task. In the qualitative results, like the quantitative results, there is a performance drop when increasing vocabulary size and including attributes. 

\section{Conclusion and Future Work}
In this paper we proposed a novel technique for learning to represent an image via a scene graph. Our approach is based on generating individual subgraphs called triples, and exploiting an attention mechanism to stitch triples together into a proper scene graph.  We hypothesized that by focusing on generating individual true statements  about a scene in the form of triples, we can create better triples and thus better scene graphs.  After triples are generated, we made explicit use of vectors from an attention mechanism in our generator to determine which lexemes in the triples refer to the same object, yielding a method that combines lexemes into a single node in our scene graph.  To our knowledge this is a novel use of attention in neural networks. By learning our model using an adversarial training technique, our method can be trained to generate object attributes in addition to object relations, which prior work did not do, and without bounding box information, which prior work required.  In an empirical evaluation, we illustrate that our method for generating scene graphs outperforms the current state-of-the-art, achieving double the recall@50 and recall@100.

One potential experiment that can be explored in future work is the insertion of various architectures into the GAN framework. For example, while our convolutional features are fixed and extracted from a CNN trained to recognize images, training a convolutional component of the network jointly with our LSTM may improve performance as the tasks of image recognition and scene graph generation are sufficiently different. Another line of research includes exploring other structured prediction problems given a complex input (like an image) using GANs. For example, while we argued for the adoption of scene graphs over image captions as image representations, image captioning has proven to be useful and could potentially benefit from an adversarial training approach. There are also unexplored lines of research in working with scene graphs themselves. Structured knowledge sources have a number of applications that have not yet found their way to using scene graphs as a source, e.g faceted search. Lastly we suggest that scene graph provide a means of linking information in an image with information from data of other modalities, for example text. Since many information extraction techniques for text produce graph structured results, they may be compatible with scene graphs.

\section{Acknowledgements}
Matt Klawonn is supported by an ASEE SMART scholarship. The authors would like to thank professor James Hendler for his thoughtful comments and discussion.


\begin{thebibliography}{1}

\bibitem[\protect\citeauthoryear{Anderson \bgroup et al\mbox.\egroup
  }{2016}]{anderson2016spice}
Anderson, P.; Fernando, B.; Johnson, M.; and Gould, S.
\newblock 2016.
\newblock Spice: Semantic propositional image caption evaluation.
\newblock {\em arXiv preprint arXiv:1607.08822}.

\bibitem[\protect\citeauthoryear{Arjovsky and
  Bottou}{2017}]{arjovsky2017towards}
Arjovsky, M., and Bottou, L.
\newblock 2017.
\newblock Towards principled methods for training generative adversarial
  networks.
\newblock {\em arXiv preprint arXiv:1701.04862}.

\bibitem[\protect\citeauthoryear{Arjovsky, Chintala, and
  Bottou}{2017}]{arjovsky2017wasserstein}
Arjovsky, M.; Chintala, S.; and Bottou, L.
\newblock 2017.
\newblock Wasserstein generative adversarial networks.

\bibitem[\protect\citeauthoryear{Ba, Kiros, and Hinton}{2016}]{ba2016layer}
Ba, J.~L.; Kiros, J.~R.; and Hinton, G.~E.
\newblock 2016.
\newblock Layer normalization.
\newblock {\em arXiv preprint arXiv:1607.06450}.

\bibitem[\protect\citeauthoryear{Bahdanau, Cho, and
  Bengio}{2014}]{bahdanau2014neural}
Bahdanau, D.; Cho, K.; and Bengio, Y.
\newblock 2014.
\newblock Neural machine translation by jointly learning to align and
  translate.
\newblock {\em arXiv preprint arXiv:1409.0473}.

\bibitem[\protect\citeauthoryear{Chung \bgroup et al\mbox.\egroup
  }{2014}]{chung2014empirical}
Chung, J.; Gulcehre, C.; Cho, K.; and Bengio, Y.
\newblock 2014.
\newblock Empirical evaluation of gated recurrent neural networks on sequence
  modeling.
\newblock {\em arXiv preprint arXiv:1412.3555}.

\bibitem[\protect\citeauthoryear{Ferrari and
  Zisserman}{2008}]{ferrari2008learning}
Ferrari, V., and Zisserman, A.
\newblock 2008.
\newblock Learning visual attributes.
\newblock In {\em Neural Information Processing Systems}.

\bibitem[\protect\citeauthoryear{Goodfellow \bgroup et al\mbox.\egroup
  }{2014}]{goodfellow2014generative}
Goodfellow, I.~J.; Pouget-Abadie, J.; Mirza, M.; Xu, B.; Warde-Farley, D.;
  Ozair, S.; Courville, A.; and Bengio, Y.
\newblock 2014.
\newblock Generative adversarial networks.
\newblock {\em arXiv preprint arXiv:1406.2661}.

\bibitem[\protect\citeauthoryear{Gulrajani \bgroup et al\mbox.\egroup
  }{2017}]{gulrajani2017improved}
Gulrajani, I.; Ahmed, F.; Arjovsky, M.; Dumoulin, V.; and Courville, A.
\newblock 2017.
\newblock Improved training of wasserstein gans.
\newblock {\em arXiv preprint arXiv:1704.00028}.

\bibitem[\protect\citeauthoryear{He \bgroup et al\mbox.\egroup
  }{2015}]{he2015deep}
He, K.; Zhang, X.; Ren, S.; and Sun, J.
\newblock 2015.
\newblock Deep residual learning for image recognition.
\newblock {\em arXiv preprint arXiv:1512.03385}.

\bibitem[\protect\citeauthoryear{Hochreiter and
  Schmidhuber}{1997}]{hochreiter1997long}
Hochreiter, S., and Schmidhuber, J.
\newblock 1997.
\newblock Long short-term memory.
\newblock {\em Neural computation} 9(8):1735--1780.

\bibitem[\protect\citeauthoryear{Jang, Gu, and
  Poole}{2016}]{jang2016categorical}
Jang, E.; Gu, S.; and Poole, B.
\newblock 2016.
\newblock Categorical reparameterization with gumbel-softmax.
\newblock {\em arXiv preprint arXiv:1611.01144}.

\bibitem[\protect\citeauthoryear{Johnson \bgroup et al\mbox.\egroup
  }{2017}]{johnson2015image}
Johnson, J.; Krishna, R.; Stark, M.; Li, L.-J.; Shamma, D.; Bernstein, M.; and
  Fei-Fei, L.
\newblock 2017.
\newblock Image retrieval using scene graphs.
\newblock In {\em Computer Vision and Pattern Recognition}.

\bibitem[\protect\citeauthoryear{Johnson, Karpathy, and
  Fei-Fei}{2015}]{johnson2015densecap}
Johnson, J.; Karpathy, A.; and Fei-Fei, L.
\newblock 2015.
\newblock Densecap: Fully convolutional localization networks for dense
  captioning.
\newblock {\em arXiv preprint arXiv:1511.07571}.

\bibitem[\protect\citeauthoryear{Kingma and Ba}{2014}]{kingma2014adam}
Kingma, D.~P., and Ba, J.
\newblock 2014.
\newblock Adam: A method for stochastic optimization.
\newblock {\em arXiv preprint arXiv:1412.6980}.

\bibitem[\protect\citeauthoryear{Krishna \bgroup et al\mbox.\egroup
  }{2016}]{krishna2016visual}
Krishna, R.; Zhu, Y.; Groth, O.; Johnson, J.; Hata, K.; Kravitz, J.; Chen, S.;
  Kalantidis, Y.; Li, L.-J.; Shamma, D.~A.; et~al.
\newblock 2016.
\newblock Visual genome: Connecting language and vision using crowdsourced
  dense image annotations.
\newblock {\em arXiv preprint arXiv:1602.07332}.

\bibitem[\protect\citeauthoryear{Krizhevsky, Sutskever, and
  Hinton}{2012}]{krizhevsky2012imagenet}
Krizhevsky, A.; Sutskever, I.; and Hinton, G.~E.
\newblock 2012.
\newblock Imagenet classification with deep convolutional neural networks.
\newblock In {\em Neural Information Processing Systems}.

\bibitem[\protect\citeauthoryear{Lassila and Swick}{1999}]{lassila1999resource}
Lassila, O., and Swick, R.~R.
\newblock 1999.
\newblock Resource description framework (rdf) model and syntax specification.

\bibitem[\protect\citeauthoryear{Ledig \bgroup et al\mbox.\egroup
  }{2016}]{ledig2016photo}
Ledig, C.; Theis, L.; Huszar, F.; Caballero, J.; Cunningham, A.; Acosta, A.;
  Aitken, A.; Tejani, A.; Totz, J.; Wang, Z.; et~al.
\newblock 2016.
\newblock Photo-realistic single image super-resolution using a generative
  adversarial network.
\newblock {\em arXiv preprint arXiv:1609.04802}.

\bibitem[\protect\citeauthoryear{Lu \bgroup et al\mbox.\egroup
  }{2016}]{lu2016hierarchical}
Lu, J.; Yang, J.; Batra, D.; and Parikh, D.
\newblock 2016.
\newblock Hierarchical question-image co-attention for visual question
  answering.
\newblock {\em arXiv preprint arXiv:1606.00061}.

\bibitem[\protect\citeauthoryear{Maddison, Mnih, and
  Teh}{2016}]{maddison2016concrete}
Maddison, C.~J.; Mnih, A.; and Teh, Y.~W.
\newblock 2016.
\newblock The concrete distribution: A continuous relaxation of discrete random
  variables.
\newblock {\em arXiv preprint arXiv:1611.00712}.

\bibitem[\protect\citeauthoryear{Malinowski and
  Fritz}{2014}]{malinowski2014towards}
Malinowski, M., and Fritz, M.
\newblock 2014.
\newblock Towards a visual turing challenge.
\newblock {\em arXiv preprint arXiv:1410.8027}.

\bibitem[\protect\citeauthoryear{Radford, Metz, and
  Chintala}{2015}]{radford2015unsupervised}
Radford, A.; Metz, L.; and Chintala, S.
\newblock 2015.
\newblock Unsupervised representation learning with deep convolutional
  generative adversarial networks.
\newblock {\em arXiv preprint arXiv:1511.06434}.

\bibitem[\protect\citeauthoryear{Reed \bgroup et al\mbox.\egroup
  }{2016}]{reed2016generative}
Reed, S.; Akata, Z.; Yan, X.; Logeswaran, L.; Schiele, B.; and Lee, H.
\newblock 2016.
\newblock Generative adversarial text to image synthesis.
\newblock {\em arXiv preprint arXiv:1605.05396}.

\bibitem[\protect\citeauthoryear{Russakovsky \bgroup et al\mbox.\egroup
  }{2014}]{russakovsky2014imagenet}
Russakovsky, O.; Deng, J.; Su, H.; Krause, J.; Satheesh, S.; Ma, S.; Huang, Z.;
  Karpathy, A.; Khosla, A.; Bernstein, M.; et~al.
\newblock 2014.
\newblock Imagenet large scale visual recognition challenge.
\newblock {\em arXiv preprint arXiv:1409.0575}.

\bibitem[\protect\citeauthoryear{Shih, Singh, and Hoiem}{2015}]{shih2015look}
Shih, K.~J.; Singh, S.; and Hoiem, D.
\newblock 2015.
\newblock Where to look: Focus regions for visual question answering.
\newblock {\em arXiv preprint arXiv:1511.07394}.

\bibitem[\protect\citeauthoryear{Simonyan and
  Zisserman}{2014}]{simonyan2014very}
Simonyan, K., and Zisserman, A.
\newblock 2014.
\newblock Very deep convolutional networks for large-scale image recognition.
\newblock {\em arXiv preprint arXiv:1409.1556}.

\bibitem[\protect\citeauthoryear{Szegedy \bgroup et al\mbox.\egroup
  }{2016}]{szegedy2016inception}
Szegedy, C.; Ioffe, S.; Vanhoucke, V.; and Alemi, A.
\newblock 2016.
\newblock Inception-v4, inception-resnet and the impact of residual connections
  on learning.
\newblock {\em arXiv preprint arXiv:1602.07261}.

\bibitem[\protect\citeauthoryear{Vinyals \bgroup et al\mbox.\egroup
  }{2014}]{vinyals2014show}
Vinyals, O.; Toshev, A.; Bengio, S.; and Erhan, D.
\newblock 2014.
\newblock Show and tell: A neural image caption generator.
\newblock {\em arXiv preprint arXiv:1411.4555}.

\bibitem[\protect\citeauthoryear{Xu and Saenko}{2015}]{xu2015ask}
Xu, H., and Saenko, K.
\newblock 2015.
\newblock Ask, attend and answer: Exploring question-guided spatial attention
  for visual question answering.
\newblock {\em arXiv preprint arXiv:1511.05234}.

\bibitem[\protect\citeauthoryear{Xu \bgroup et al\mbox.\egroup
  }{2015}]{xu2015show}
Xu, K.; Ba, J.; Kiros, R.; Cho, K.; Courville, A.; Salakhutdinov, R.; Zemel,
  R.; and Bengio, Y.
\newblock 2015.
\newblock Show, attend and tell: Neural image caption generation with visual
  attention.
\newblock {\em arXiv preprint arXiv:1502.03044}.

\bibitem[\protect\citeauthoryear{Xu \bgroup et al\mbox.\egroup
  }{2017}]{xu2017scenegraph}
Xu, D.; Zhu, Y.; Choy, C.~B.; and Fei-Fei, L.
\newblock 2017.
\newblock Scene graph generation by iterative message passing.
\newblock {\em arXiv preprint arXiv:1701.02426}.

\bibitem[\protect\citeauthoryear{Yang \bgroup et al\mbox.\egroup
  }{2015}]{yang2015stacked}
Yang, Z.; He, X.; Gao, J.; Deng, L.; and Smola, A.
\newblock 2015.
\newblock Stacked attention networks for image question answering.
\newblock {\em arXiv preprint arXiv:1511.02274}.

\bibitem[\protect\citeauthoryear{Yu \bgroup et al\mbox.\egroup
  }{2016}]{yu2016sequence}
Yu, L.; Zhang, W.; Wang, J.; and Seqgan, Y.~Y.
\newblock 2016.
\newblock sequence generative adversarial nets with policy gradient. arxiv
  preprint.
\newblock {\em arXiv preprint arXiv:1609.05473} 2(3):5.

\bibitem[\protect\citeauthoryear{Zhang \bgroup et al\mbox.\egroup
  }{2017}]{han2017stackgan}
Zhang, H.; Xu, T.; Li, H.; Zhang, S.; Wang, X.; Huang, X.; and Metaxas, D.
\newblock 2017.
\newblock Stackgan: Text to photo-realistic image synthesis with stacked
  generative adversarial networks.
\newblock In {\em International Conference on Computer Vision}.

\end{thebibliography}
\end{document}